%% file: neurips_2022.tex
\documentclass{article}

\PassOptionsToPackage{numbers, compress}{natbib}

\usepackage[preprint]{neurips_2022}

\usepackage[dvipsnames]{xcolor}
\definecolor{bleublue}{rgb}{0.3, 0.5, 0.9}
\definecolor{caribbeangreen}{rgb}{0.5, 0.9, 0.6}
\usepackage[utf8]{inputenc} 
\usepackage[T1]{fontenc}    
\usepackage{hyperref}       
\usepackage{url}            
\usepackage{booktabs}       
\usepackage{amsfonts}       
\usepackage{nicefrac}       
\usepackage{microtype}      
\usepackage[dvipsnames]{xcolor}         
\usepackage{graphicx}       
\usepackage{subcaption}
\usepackage{wrapfig}
\usepackage[export]{adjustbox}
\usepackage{amsmath}

\input{tex/macros.tex}

\title{UNeRF: Time and Memory Conscious U-Shaped Network for Training Neural Radiance Fields}

\author{Abiramy Kuganesan \hspace{5mm} Shih-yang Su \hspace{5mm} James J. Little \hspace{5mm} Helge Rhodin \\
University of British Columbia \\
{\tt\small \{akuganes, shihyang, little, rhodin\}@cs.ubc.ca}}

\begin{document}

\maketitle
\input{tex/abstract}
\input{tex/intro}
\input{tex/relatedwork}
\input{tex/nerfprelim}
\input{tex/method}
\input{tex/evaluation}

\input{tex/conclusion}

\begin{ack}
This work was sponsored in part by the Natural Sciences and Engineering Research Council of Canada. We thank Compute Canada for access to their GPU servers.
\end{ack}

\section*{References}
{
\renewcommand{\bibsection}{}
\bibliographystyle{splncs04}
\medskip
\bibliography{bib}
}

\end{document}

%% file: tex/macros.tex
\usepackage{soul} 

\newcommand{\parag}[1]{{\bf{#1}\quad}}

\newcommand{\R}{\mathbb{R}}

\newcommand{\vc}{\mathbf{c}}
\newcommand{\vd}{\mathbf{d}}

\newcommand{\vr}{\mathbf{r}}

\newcommand{\vx}{\mathbf{x}}

%% file: tex/abstract.tex
\begin{abstract}

Neural Radiance Fields (NeRFs) increase reconstruction detail for novel view synthesis and scene reconstruction, with applications ranging from large static scenes to dynamic human motion. However, the increased resolution and model-free nature of such neural fields come at the cost of high training times and excessive memory requirements. Recent advances improve the inference time by using complementary data structures yet these methods are ill-suited for dynamic scenes and often increase memory consumption. Little has been done to reduce the resources required at training time. We propose a method to exploit the redundancy of NeRF's sample-based computations by partially sharing evaluations across neighboring sample points. Our UNeRF architecture is inspired by the UNet, where spatial resolution is reduced in the middle of the network and information is shared between adjacent samples. Although this change violates the strict and conscious separation of view-dependent appearance and view-independent density estimation in the NeRF method, we show that it improves novel view synthesis. We also introduce an alternative subsampling strategy which shares computation while minimizing any violation of view invariance. UNeRF is a plug-in module for the original NeRF network. Our major contributions include reduction of the memory footprint, improved accuracy, and reduced amortized processing time both during training and inference. With only weak assumptions on locality, we achieve improved resource utilization on a variety of neural radiance fields tasks. We demonstrate applications to the novel view synthesis of static scenes as well as dynamic human shape and motion.
\end{abstract}

%% file: tex/intro.tex
\section{Introduction}
\input{figures/fig_differencemap}

A key stepping stone in the field of model-free reconstruction was the introduction of neural radiance fields (NeRF) \cite{mildenhall2020nerf} for rendering novel photorealistic views from static images. Such neural network-driven scene encoding succeeds without spatial discretization, thereby avoiding the excessive memory requirement of voxel grids and the difficulty of optimizing topological changes for meshes. However, the underlying neural network still has a large memory footprint and requires thousands of evaluations per image. For complex scenes, the process of training a NeRF model easily spans days on multiple high-end GPUs, making experimentation inaccessible for practitioners with limited resources and imposing colossal environmental damage when applied at scale.

There exists promising work on improving the runtime of neural radiance fields by using voxel grids~\cite{liu2020nsvf}, Voronoi partitions~\cite{Rebain2021DeRFDR}, or entirely replacing the neural network with a tree~\cite{yu2021plenoctrees}. These acceleration structures are mostly targeted to improve inference time and may even add on to the training time. A very promising recent approach is to use hashing~\cite{mueller2022instant} which improves both training and test time at the expense of massive memory consumption during training time. Moreover, all of these methods are designed for static scenes making them inapplicable for dynamic scenes such as human motion.

Our goal is to replace the multilayer perceptron (MLP) used in NeRF and many other neural field approaches~\cite{nerfinthewild,park2021hypernerf,xie2021neural} with a more efficient network that exploits local continuity while not requiring restrictive conditions such as the assumption of a static scene. NeRF employs an MLP to represent the scene as a radiance field that stores the amount of light emitted at every point and direction in space. This is then paired with a density field, computed by a branch of the same MLP, to model opacity. An image is formed by classical ray-marching~\cite{classicalraymarching} by casting a ray from the camera through every pixel and integrating the light emitted towards the camera along discrete samples while accounting for occlusions using the density network. The costly component is the evaluation of the MLP for every step along the ray, with the number of total samples usually in the order of hundreds of thousands, especially when importance sampling is applied.

Our underlying idea is to share computations between neighboring samples, building upon the fact that only high-frequency details should change at this capacity. The challenge is to create a structure that enables the network to learn what can and cannot be shared while maintaining beneficial view-invariances where possible. We take inspiration from the UNet~\cite{ronneberger2015unet} architecture used for 2D image segmentation which gradually reduces the spatial resolution in the first half of the network via convolution and subsequently upsamples the resulting high-level features in the second half with the use of skip connections and deconvolution. In the same vein, we apply convolutions along individual 1D rays to share information between neighboring scene points. It is worthy of note that this violates the consciously chosen view-direction invariances in NeRF, as combining the features of two sample points alludes to the ray direction between them. However, we show that the benefits in model expressiveness outweigh this sacrifice, resulting in improved image quality which suggests that these invariances can also be learned if sufficient training data is available. In addition, we also provide a variant with subsampling which preserves invariances by limiting cues on the ray direction while forming the same U-shaped network structure. The latter reduces training time, inference time, and memory the footprint but comes at the cost of a slight loss in reconstruction accuracy, while the former improves all properties including optimization time.

Unlike up-convolutions~\cite{odena2016deconvolution} or bi-linear interpolation in UNet~\cite{ronneberger2015unet} and related convolutional neural networks~\cite{long2015fully,Aitken2017CheckerboardAF,dcgan2015radford}, samples along the ray for NeRF are not uniformly spaced due to importance sampling. Our additional technical contribution is that we improve by linearly interpolating based on the query location instead of querying against a regular grid.

We experiment with different instantiations of the UNeRF's down-sampling and up-scaling strategies, showing that the proposed network consistently outperforms NeRF and other simpler baselines on the most widely used benchmarks. UNeRF conserves 21\% of the memory consumed by NeRF while cutting down training time by 12\%, and exhibiting better perceptual quality than NeRF. By applying it to learning human body shape from videos~\cite{su2021anerf}, we further demonstrate the versatility and plug-in nature of our approach, including its application to dynamic scenes.

%% file: figures/fig_differencemap.tex
\begin{figure}[]
\begin{subfigure}[t]{\textwidth}
\minipage{0.13\textwidth}
  \includegraphics[width=\linewidth, trim={0 0 0 0},clip]{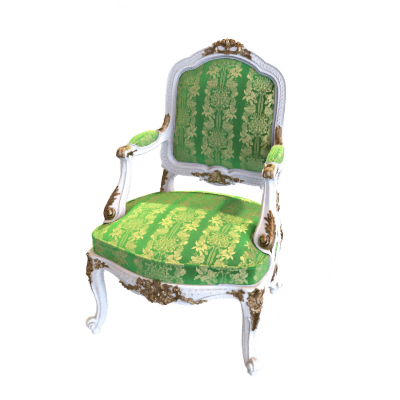}
\endminipage\hfill
\minipage{0.15\textwidth}
  \includegraphics[width=\linewidth, trim={0 0 0 0},clip]{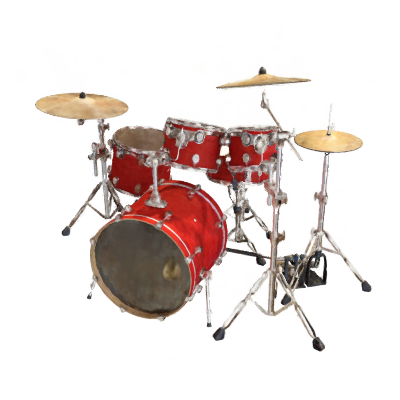}
\endminipage\hfill
\minipage{0.13\textwidth}
  \includegraphics[width=\linewidth, trim={0 0 0 0},clip]{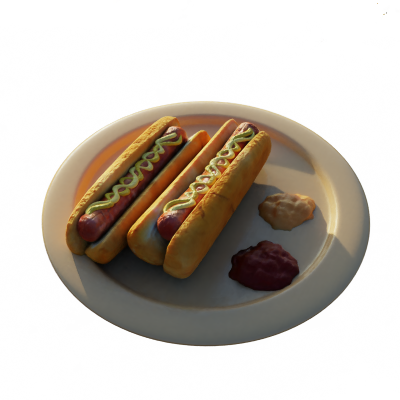}
\endminipage\hfill
\minipage{0.13\textwidth}
  \includegraphics[width=\linewidth, trim={0 0 0 0},clip]{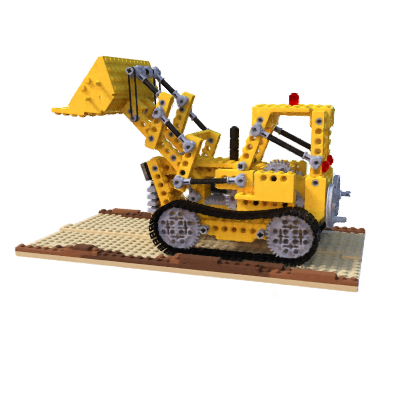}
\endminipage\hfill
\minipage{0.13\textwidth}
  \includegraphics[width=\linewidth, trim={0 0 0 0},clip]{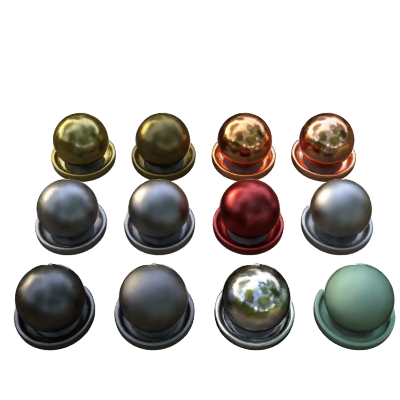}
\endminipage\hfill
\minipage{0.13\textwidth}
  \includegraphics[width=\linewidth, trim={0 0 0 0},clip]{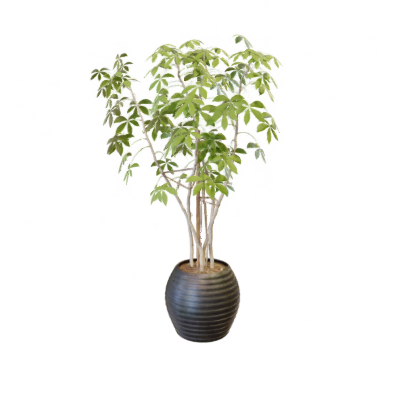}
\endminipage\hfill
\minipage{0.13\textwidth}%
  \includegraphics[width=\linewidth, trim={0 0.5cm 0 0},clip]{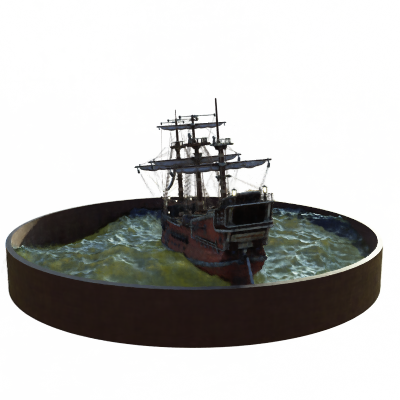}
\endminipage
\end{subfigure}
\begin{subfigure}[t]{\textwidth}
\minipage{0.13\textwidth}
  \includegraphics[width=\linewidth, trim={0 0 0 0},clip]{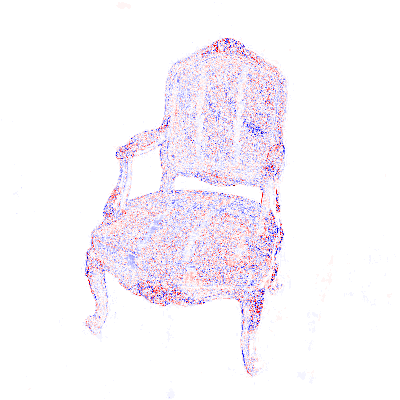}
\endminipage\hfill
\minipage{0.15\textwidth}
  \includegraphics[width=\linewidth, trim={0 0 0 0},clip]{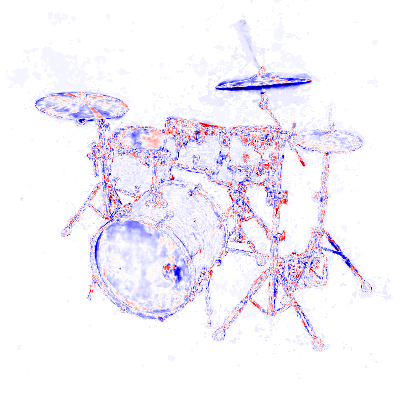}
\endminipage\hfill
\minipage{0.13\textwidth}
  \includegraphics[width=\linewidth, trim={0 0 0 0},clip]{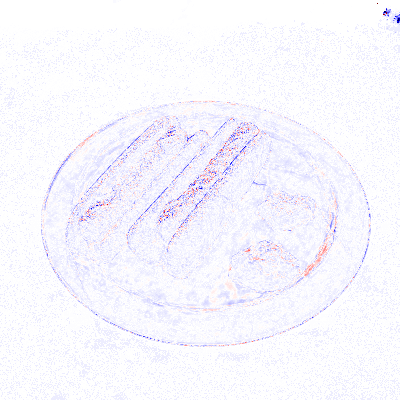}
\endminipage\hfill
\minipage{0.13\textwidth}
  \includegraphics[width=\linewidth, trim={0 0 0 0},clip]{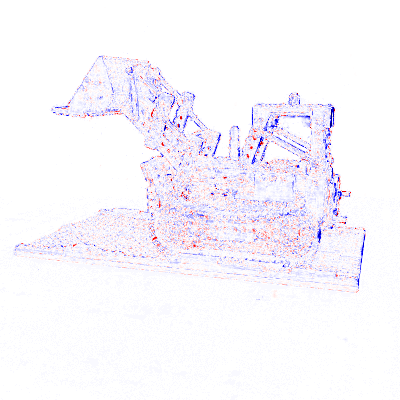}
\endminipage\hfill
\minipage{0.13\textwidth}
  \includegraphics[width=\linewidth, trim={0 0 0 0},clip]{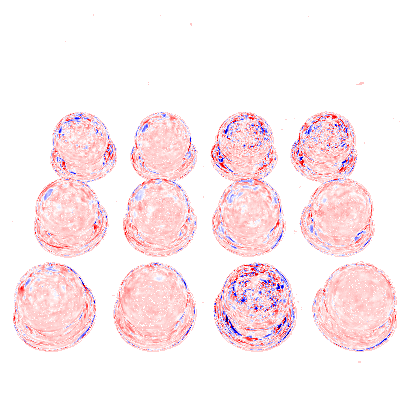}
\endminipage\hfill
\minipage{0.13\textwidth}
  \includegraphics[width=\linewidth, trim={0 0 0 0},clip]{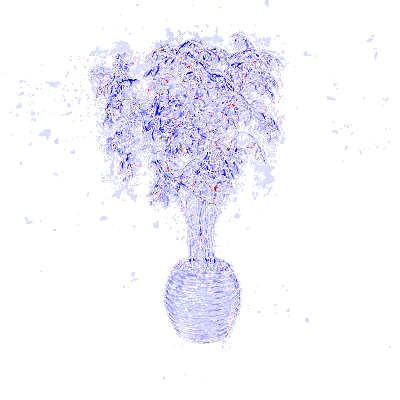}
\endminipage\hfill
\minipage{0.13\textwidth}%
  \includegraphics[width=\linewidth, trim={0 0 0 0},clip]{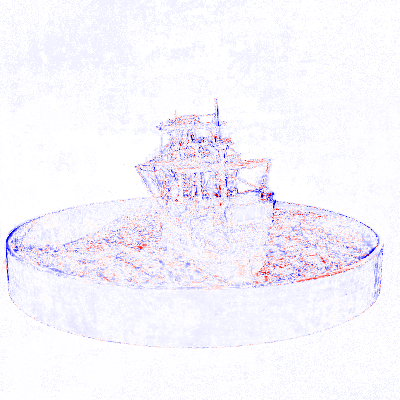}
\endminipage
\end{subfigure}
\begin{subfigure}[t]{\textwidth}
    \includegraphics[width=\linewidth, trim={0 1.7cm 0 0},clip]{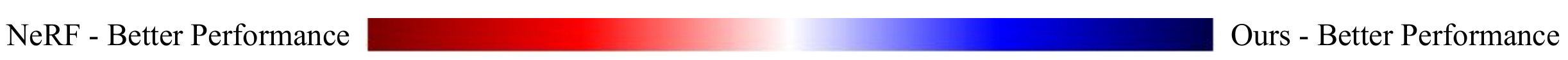}
\end{subfigure}
\caption{\textbf{Qualitative comparisons on the Blender dataset.} Top row: Rendered testset images from UNeRF-Conv with a $3\times1$ kernel. Bottom row: Difference maps between UNeRF-Conv, NeRF and the ground truth. Pixels highlighted more prominently in {\color{blue}blue} indicate regions where UNeRF-Conv more closely aligns with the ground truth than NeRF, and {\color{red}red} the vice versa. The pixel color difference is normalized to [0, 1]. While NeRF and UNeRF-Conv typically exhibits comparable results when describing low frequency details, UNeRF-Conv has an edge on depicting high frequency detail more accurately.}
\label{fig: differencemaps}
\end{figure}

%% file: tex/relatedwork.tex
\section{Related work}

In the following section, we review the algorithmic advances and trade-offs in the literature for improving NeRF's imminent resource demand and put them in contrast to our proposed method which builds upon NeRF.

\paragraph{Discrete grids and acceleration aids}
Explicit volumetric grids have the advantage of a regular structure that can be traversed by raymarching efficiently to significantly improve training \cite{liu2020nsvf,Sun2021DirectVoxelGrids} and inference time
\cite{Lombardi2019NeuralVolumes,Sai2020DeepVolumes}. However, 3D grids have cubic memory consumption limiting their resolution to about $128^3$ on current hardware \cite{Lombardi2019NeuralVolumes,Sai2020DeepVolumes}. Classical acceleration structures such as octrees reduce computation by avoiding sampling empty space, either by directly storing radiance~\cite{Garbin2021FastNeRF,Wizadwongsa2021NeX,yu2021plenoctrees} or by subdividing the space that needs to be sampled~\cite{liu2020nsvf,takikawa2021nglod}.
The former was taken to the extreme with a hierarchical hash table of training feature vectors \cite{mueller2022instant} that can train in seconds and render in milliseconds by reducing the network to two fully connected layers to disambiguate hash collisions. However, the speed comes at the expense of an exceptional memory burden during training time.
DeRF \cite{Rebain2021DeRFDR} chooses an alternative decomposition into Voronoi cells, yielding 3x rendering speedup with improved perceptual quality scores. However, while introducing a rendering speedup, these methods require a separate construction step, often using a conversion precursor from a trained NeRF~\cite{yu2021plenoctrees,Garbin2021FastNeRF,hedman2021baking}, thus experiencing lengthier training times. Moreover, the underlying acceleration structures that partition the 3D scene statically are not applicable to the dynamic scenes we target. 

\paragraph{Importance sampling and integration}
A key ingredient in NeRF is the concept of importance sampling, which clusters samples along visible surfaces with a coarse-to-fine network. The number of samples can be further reduced by depth supervision \cite{kangle2021dsnerf,Piala2021TermiNeRF} or through learning to predict a depth map~\cite{neff2021donerf}, improving rendering time to 15 FPS at the cost of half of the visual quality of NeRF. AutoInt \cite{lindell2021autoint} substitutes Monte Carlo sampling to a learned integration function which succeeds with as few as two evaluations along each ray. While providing training and render-time speedups, the image quality reduces to 86\% of a NeRF with the same capacity. By contrast, we keep the original NeRF sampling strategy but partially share computations among neighboring samples which is shown to outperform NeRF and this baseline.

\paragraph{Architectural adaptations}
Work centered around explicit volumetric representations have attempted to optimize the complex MLP for a reduction in rendering time \cite{reiser2021kilonerf,Garbin2021FastNeRF,yu2021plenoctrees}. Opting for compiled languages~\cite{yu2021plenoxels} instead of interpreted ones as well as parallelization across multiple processors~\cite{jaxnerf2020github} have shown promise in reducing training time. KiloNeRF \cite{reiser2021kilonerf} factorizes the MLP into a large set of very small MLPs.  FastNeRF \cite{Garbin2021FastNeRF} restructures the MLP into two MLPs supported by caching. We relax the fundamental assumption that neighboring samples must be independent and that density estimates are strictly view-direction independent, which enables more efficient network architectures that apply to static and dynamic scenes. IBRNet~\cite{wang2021ibrnet} is an image-based rendering approach that uses a transformer to integrate information from multiple views for novel-view synthesis. In doing so, they exploit view dependence for improved accuracy.

The main obstacle with current methods is the trade-off between compute time, memory, and perceptual quality especially during training. We seek a robust method which reduces optimization time, expresses lower memory consumption and results in improved perceptual image quality--- one that is as task agnostic as the original NeRF method. 

%% file: tex/nerfprelim.tex
\section{NeRF preliminaries} Given a position $\vx \in \R^3$ and direction $\vd \in \R^3$ as input, NeRF~\cite{mildenhall2020nerf} predicts direction-independent density $\sigma(\vx) \in \R$ and direction-dependent color $\vc(\vx,\vd) \in \R^3$.
Such scene depiction is rendered by casting rays from a virtual camera, sampling $N$ infinitesimal points $\vx_i$ along each ray $\vr$, evaluating the model at these points, and integrating the radiance along the ray while accounting for transmission through the density field. The core design choice of the original NeRF implementation is that $\sigma(\vx)$ is made independent of view direction $\vd$ with an implicit bias. By conditioning on each position $\vx_1,\dots,\vx_N$ independently and first predicting $\sigma(\vx)$ with a sequence of MLPs before conditioning on $\vd$, NeRF is able to predict $\vc(\vx_i,\vd)$ for the i-th sample with just two additional layers. This design was reinforced with positional encoding~\cite{vaswani2017positionalencoding}, where projecting $\vx_i$ onto a series of sine and cosine waves of increasing frequency $\gamma(\vx_i)$ thereby enables the model to attend to low frequencies where possible and to high frequencies where necessary~\cite{vaswani2017positionalencoding,tancik2020fourfeat}. The advantage is that this neural representation is differentiable, lending itself to the end-to-end reconstruction of 3D scenes from raw images. Importance sampling is employed to query primarily those points along the ray that contribute to the visible surface, which makes the learning of high-resolution models tractable. A coarse network evaluates uniformly distributed samples to inform an identical fine network with additional non-uniform importance samples. Even when using importance sampling to limit the number of randomly distributed uniform samples, the main disadvantages are that detailed scenes require hundreds of samples with corresponding network invocations for rendering a single pixel and typical NeRF networks are deep with up to 10 layers, each with 256 channels. 

%% file: tex/method.tex
\section{Method}
\label{sec: method}
\input{figures/unerf_architecture}

We aim to improve the efficiency of NeRF by sharing some computations between adjacent samples along the ray direction since samples in close proximity share smoothly changing field properties. To this end, we propose two variants of UNeRF which both reduce resource consumption by downsampling the spatial resolution of parts of the network. The first variant is an adaptation of the UNet~\cite{ronneberger2015unet} for a 1D convolutional configuration along the samples of a ray. It departs from NeRF's use of implicit bias to ensure view invariances, thereby opening the door for data-driven learning of dependencies between neighboring samples. The latter uses subsampling instead of convolutions to reduce the resolution of feature vectors through shared computation while maintaining view invariances. Both variants utilize position-aware interpolation to account for the non-uniform sampling structure in comparison to the regular 2D grids in the original UNet application.

\subsection{UNeRF-Conv}
\label{sec: unerfconv}
Starting from the original 10-layer MLP architecture used by NeRF, we replace the six middle layers with a variant of the 2D UNet adapted to 1D. Instead of processing sample points $\vx_1,\dots,\vx_N$ independently as in NeRF, information between neighboring features is exchanged with convolutions along the ray direction. Layers 2, 3, and 4 operate at a reduced spatial resolution by employing a $k\times1$ kernel with a stride of size 2. Each of these convolutions is followed by a ReLU activation function to mimic the functional equivalent of a fully connected layer, coupled with a subsampling operation, followed by a ReLU activation function. Subsequently, the outputs of layers 4, 5, and 6 are upsampled with position-aware linear interpolation, creating the U shape in Figure~\ref{fig: unerfarchitecture}. These up-scaling layers are fully connected followed by a ReLU activation function. Borrowing inspiration from ResNet~\cite{He2016resnet}, we make use of skip connections which help pass high-frequency information through the spatial bottleneck. This mitigates the otherwise common bias of MLPs towards lower frequency features \cite{rahaman2019biasToLowFreq} and ensures that computations can be shared across samples-- increasing the model capacity without consuming additional memory.

Neighboring points along the ray involuntarily encode the view direction since samples are exclusively chosen along these rays during training and inference. The restrictive $k$-dimensional window of the 1D convolutions and irregular sampling limit its violation of the implicit bias on view invariance, allowing the 3D shape and appearance to be reconstructed using this UNeRF-Conv variant.

\subsection{UNeRF-Sub}
\label{sec: unerfsub}

\input{figures/fig_FeatureVectorManipulation}

In order to share information while minimizing the violation of view invariance, we propose UNeRF-Sub, which applies subsampling instead of the strided convolution in UNeRF-Conv. The subsampling operation is followed by the MLP's fully connected layer and subsequent ReLU activation function at layers 2, 3, and 4. Characterized by their scalar distances along the ray from the origin, points which are passed through the network are referred to as \textit{anchor points} $x'$ while points that are dropped during subsampling are referred to as \textit{intermediate points} $x''$. Figure~\ref{fig: subsampled-featureset} explains how this subsampling reduces the computational burden and memory footprint while the subsequent upsampling between anchor values with linear interpolations shares information (see Figure~\ref{fig: interpolated-featureset}). Following layer 4, we seek an incremental strategy to bring the feature resolution back to its original resolution. 

\subsection{Position-aware linear interpolation}
\label{sec: positionaware}
\input{figures/fig_InterpolationExplained}
While pixels in the UNet method are inherently
uniform, NeRF samples are chosen at irregular
intervals due to importance sampling. Hence, upsampling the latter half of UNeRF as shown in Figure~\ref{fig: unerfarchitecture} by a simple average neglects the relative positioning of samples.
Points along a ray can be parameterized by their scalar distances $x$ from the origin of the ray and feature maps $f(x)$. Given an intermediate point $x_1$, its neighbouring anchor points $x_0$ and $x_2$, and their corresponding feature vectors $f(x_0)$ and $f(x_2)$, our interpolation function $I(x_1)$ performs a weighted average on these anchor feature vectors to aprroximate the intermediate feature vector $\hat{f}(x_1)$ (see Figure \ref{fig:InterpolationExplained}).

\begin{equation}
    I(x_1) = \hat{f}(x_1) \approx f(x_0) + \frac{(f(x_2) - f(x_0))(x_1 - x_0)}{x_2 - x_0}.
    \label{eq: weightedaverage}
\end{equation}

Interleaving the interpolated feature vector set $\hat{f}(x'')$ with the anchor feature vectors $f(x')$ results in a feature set $\hat{f}(x)$ of double the resolution. This interpolation operation is visualized in Figure~\ref{fig: interpolated-featureset} and its application is shown in Figure~\ref{fig: unerfarchitecture}. Simpler interpolation strategies are discussed in Section \ref{sec: ablationstudies}.

%% file: figures/unerf_architecture.tex
\begin{figure}[t!]
  \centering
  \includegraphics[keepaspectratio, width=\textwidth, trim={0cm 0cm 0.15cm 0.75cm}, clip]{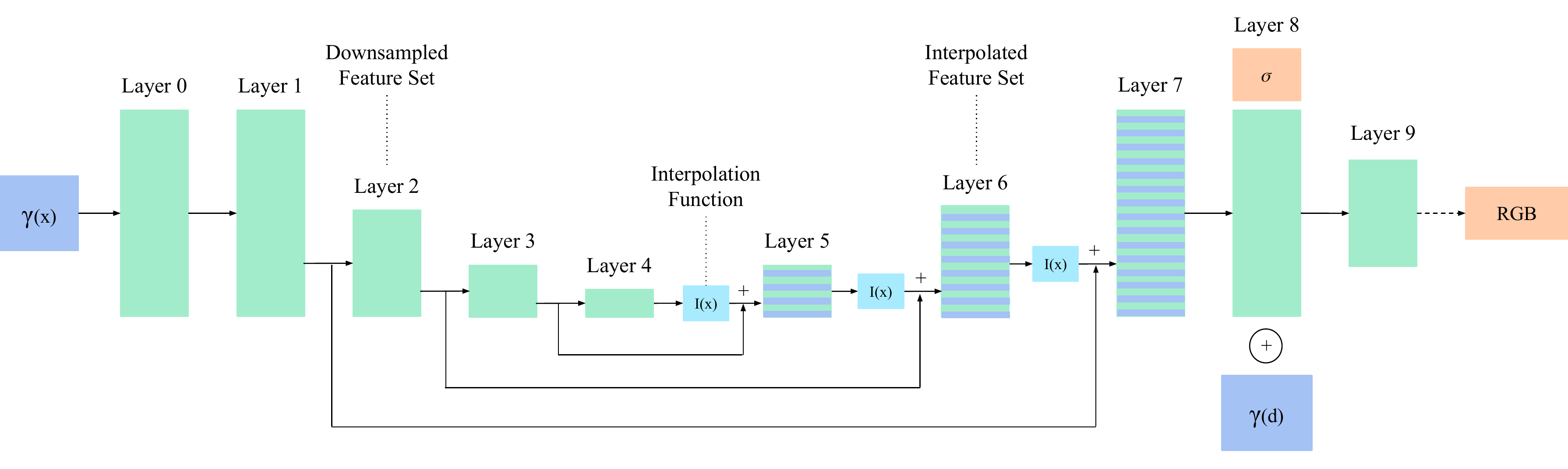}
  \caption{\textbf{UNeRF architecture.} 
  UNeRF improves computational and memory efficiency by sharing feature computations between adjacent 3D samples. Starting with positional encoded sample points, the first 5 layers process and output the feature vector with progressively smaller spatial resolution (Sections~\ref{sec: unerfconv} and \ref{sec: unerfsub}). We then apply position-aware linear interpolation (Section~\ref{sec: positionaware}) to upsample and add the interpolated features to the earlier ones to recover the feature in full spatial resolution, analogous to UNet~\cite{ronneberger2015unet}.}
  \label{fig: unerfarchitecture}
\end{figure}

%% file: figures/fig_featurevectormanipulation.tex
\begin{figure}
    \centering
    \begin{subfigure}[b]{0.32\textwidth}
        \includegraphics[width=\textwidth,trim={0.1cm 0 0 0cm}, clip]{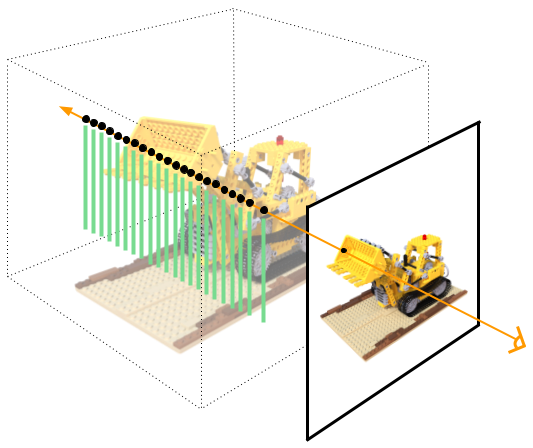}
        \caption{Full Feature Vector}
        \label{fig: full-featureset}
    \end{subfigure}
    \hfill
    \begin{subfigure}[b]{0.32\textwidth}
        \includegraphics[width=\textwidth, trim={0.1cm 0 0 0cm}, clip]{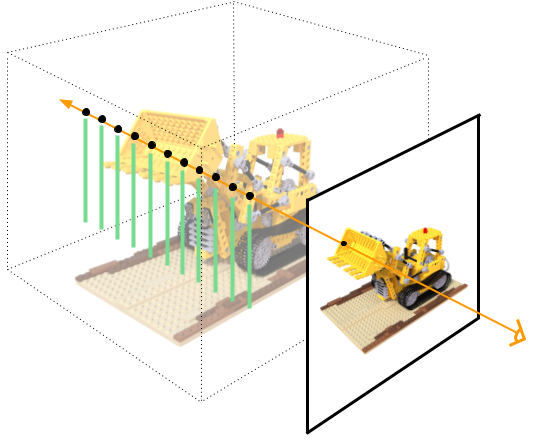}
        \caption{Downsampled Feature Vector}
        \label{fig: subsampled-featureset}
    \end{subfigure}
    \hfill
    \begin{subfigure}[b]{0.32\textwidth}
        \includegraphics[width=\textwidth, trim={0.1cm 0 0 0cm}, clip]{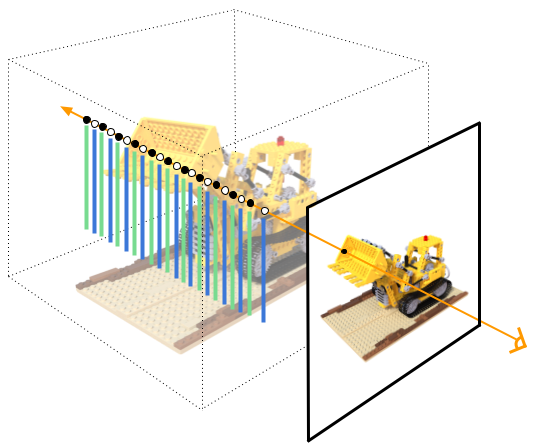}
        \caption{Interpolated Feature Vector}
        \label{fig: interpolated-featureset}
    \end{subfigure}
    \caption{\textbf{Feature vector manipulation.} Layers 0 and 1 process the features (Figure~\ref{fig: full-featureset}) corresponding to each sample along the ray. Downsampling using convolutions (Section~\ref{sec: unerfconv}) or subsampling (Section ~\ref{sec: unerfsub}) results in a downsampled feature set (Figure~\ref{fig: subsampled-featureset}) that corresponds to a sparser set of samples. Position-aware linear interpolation (Section~\ref{sec: positionaware}) brings the feature set back to its original resolution through iterative application. The interpolated feature set (Figure~\ref{fig: interpolated-featureset}) accounts for the features for samples that were omitted from MLP invocations.}
    \label{fig: featuremanipulation}
\end{figure}

%% file: figures/fig_InterpolationExplained.tex
\begin{wrapfigure}{r}{0.43\textwidth}
\vspace{-0.5cm}
    \includegraphics[width=0.43\textwidth, trim={2cm 0 3.5cm 0.5cm},clip]{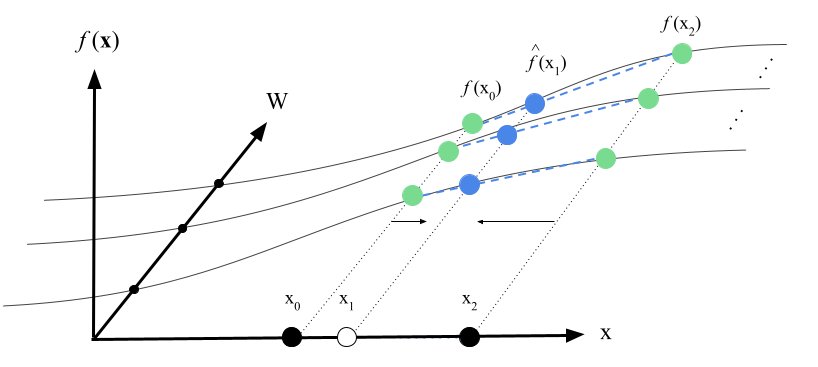}
  \caption{\textbf{{\color{bleublue}Intermediate feature vectors}} are interpolated using a linear approximation between the two adjacent \textbf{{\color{caribbeangreen}anchor feature vectors}} queried at the intermediate point.}
  \label{fig:InterpolationExplained} 
\end{wrapfigure}

%% file: tex/evaluation.tex
\section{Evaluation}
\label{sec: evaluation}
In this section, we provide an overview of the baselines and benchmarks against which we evaluate UNeRF. We perform experiments to show that our method is applicable to both dynamic and static scenes for 3D scene reconstruction. Additional experiments demonstrate UNeRF's ease of substitution for NeRF in other downstream tasks such as human motion reconstruction while exhibiting a smaller memory footprint, reduced optimization time, and better performance than NeRF itself. Ablation studies justify the various design choices which are discussed in Section~\ref{sec: method}. We encourage readers to refer to the supplementary material for additional experiments and ablation studies.

\subsection{Datasets, metrics, and baselines}
\label{sec: datasets}

\parag{Blender} We follow suit in evaluating on 7 of the static synthetic Blender scenes presented by NeRF~\cite{mildenhall2020nerf} where each synthetic object is processed at a resolution of $512 \times 512$. These 3D renderings include the Chair, Drums, Ficus, Lego, Materials, and Ship datasets. We employ an identical train, validation, and test split to the NeRF paper~\cite{mildenhall2020nerf}.

\parag{LLFF} A total of 5 complex forward facing camera scenes made available by the Local Light Field Fusion paper~\cite{mildenhallllff2019} are used. These static scenes include the Fern, Horns, Room, Trex, and Flower datasets. Each dataset ranges from 20 to 62 images. An eighth of these $1008 \times 756$ handheld phone captures are used as the test split. The data split for this dataset also follows from NeRF~\cite{mildenhallllff2019,mildenhall2020nerf}.

\parag{SURREAL} This dynamic synthetic dataset encompasses 10,800 training images from 1,200 3D human body poses rendered from 9 cameras. The testset is composed of 300 body poses from 5 different scene cameras which amounts to 1,500 test images. Although this dataset was first presented by the Learning from Synthetic Humans paper~\cite{surreal2017}, it was later adapted by A-NeRF~\cite{su2021anerf}.

\parag{Metrics} We report the peak signal to noise ratio (PSNR), the AlexNet~\cite{alexnet2012} learned perceptual image patch similarity (LPIPS)~\cite{LPIPS}, and the structural similarity index measure (SSIM)~\cite{ssim} to quantify visual image quality. For results that pertain to A-NeRF experiments, we also present the PSNR and SSIM values for the foreground to quantify the margin our method achieves with respect to the baseline on the subject of interest. In some cases, we report mean PSNR values across similar classes of datasets to summarize model performance. The maximum memory allocated during training and the median memory consumed during the forward pass are used to quantify the memory footprint. The primary time metric of interest is the amount of time it takes for a model to reach NeRF quality in terms of a fixed validation PSNR on an NVIDIA V100 GPU. We also document the median time per iteration where relevant on the same GPU model. 

\parag{Comparable models and baseline}
The default configuration suggested by NeRF~\cite{mildenhall2020nerf} is an 8 layer network with 64 coarse samples, 128 importance samples, and 4096 rays. As our objective is to present a method that optimizes memory, training time, and visual performance but all quantities influence one other, we compare models where the forward pass consumes a constant 9.06GB of GPU memory usage to enable direct comparisons. We fine-tune the number of importance samples to meet the memory target for the NeRF baseline and our UNeRF variants as our preliminary ablations (see supplemental material) demonstrate that this reduces visual fidelity the least compared to varying layer depth or width. The best-performing model variants are:
\begin{itemize}
    \item NeRF-Mem: An ablation of the NeRF architecture with only 78 importance samples.
    \item UNeRF-Fine-Sub-Mem: Our UNeRF-Sub architecture applied to only the fine network with 112 importance samples.
    \item UNeRF-Conv: Our UNeRF-Conv architecture applied to both the coarse and fine networks. 
\end{itemize}

\subsection{Object, scene, and human motion reconstruction}
\input{tables/tab_summary_synthetic_static_metrics}
\paragraph{Static synthetic scenes} The NeRF-Mem baseline using fewer importance samples converges quicker but degrades the mean visual quality. Our memory limited UNeRF-Fine-Sub-Mem network improves over this straight-forward baseline achieving 99.68\% of NeRF's visual performance at full capacity. It struggles to achieve higher resolution results beyond a fixed threshold on volumetrically dense synthetic scenes due to the compression caused by subsampling. Furthermore, UNeRF-Conv can match the full capacity NeRF baseline while consuming 24\% less memory per forward pass, 21\% less overall memory, and taking 3.86 fewer hours to converge to NeRF performance on the Chair dataset. Fully trained, the UNeRF-Conv method even outperforms NeRF by 0.93dB across 7 Blender scenes after 300k iterations.
Our method achieves better performance in all 3 categories, showing that the benefits of shared computation and using convolutions outweigh the benefits of strict view-invariances in NeRF.  
Table~\ref{tab: blender-overview-metrics} provides a quantitative overview of the results and Figure~\ref{fig: differencemaps} supplies visual comparisons of the improved reconstruction quality, especially for fine details.

\parag{Static real forward facing scenes}
UNeRF-Conv struggles against the full capacity baseline, NeRF-Mem, and UNeRF-Fine-Sub-Mem on real forward facing static scenes. We anticipate that this is due to the volumetric sparsity of the scenes in this dataset and the added difficulty in disentangling geometry from view when there is a lack of diversity in the data. The LLFF dataset captures a few images of each scene, all from only frontal views, limiting data diversity and making UNeRF-Conv unideal for these circumstances. Our UNeRF-Fine-Sub-Mem variant thrives under these conditions, achieving comparable visual performance to the full capacity NeRF with a time savings of 17\% and the aforementioned memory savings. See the supplementary material for further results and discussion. 

\input{figures/anerf}
\input{tables/tab_summary_anerf.tex}

\parag{Dynamic human motion}
The following experiments investigate UNeRF's applicability to dynamic scenes capturing non-rigid human motion. Application to the dynamic human motion problem is a critical one; most prior methods~\cite{Rebain2021DeRFDR,Sun2021DirectVoxelGrids,yu2021plenoxels} with interest in training resource optimization have not been applicable to dynamic scenes. The hyperparameter settings for these experiments come directly from the A-NeRF recommendation for the SURREAL dataset~\cite{su2021anerf}, which demonstrates our plug-in nature. The models are configured with 2048 rays, 64 coarse samples, and 16 importance samples. For evaluation, we use human poses and camera positions that are unseen during training. We compare the performance of the full NeRF model to UNeRF-Conv and UNeRF-Sub which both exhibit comparable maximum memory expenditure (see Table~\ref{table: anerf-overview-metrics}). UNeRF-Sub applied to both the coarse and fine networks suffers from aliasing as can be seen in the fingers of the subject in Figure~\ref{fig:anerf-unerfdrop}. With a maximum memory consumption of 5.24GB in contrast to 6.27GB, UNeRF-Conv consumes 20\% less memory than NeRF while exhibiting stronger foreground and overall PSNR and SSIM performance across 150k runs. This enforces that the benefit seen for static scenes translates directly to dynamic ones, which is in contrast to most existing approaches that improve efficiency only for static scenes.

\input{tables/tab_ablation_conv}

\subsection{Ablation studies}
\label{sec: ablationstudies}
\paragraph{Convolution size} Theoretically, UNeRF-Conv with kernel size $1\times1$ and stride 2 is the functional equivalent of UNeRF-Sub where every other sample is dropped during the subsampling operation. As the 1D convolution doubles as a fully connected layer and subsampling operation, when trailed by a ReLU activation function, the UNeRF-Conv $1\times1$ is the equivalent of the UNeRF-Sub implementation. As such, it makes sense that UNeRF-Sub and UNeRF Conv $1\times1$ match in PSNR performance with some noise. We perform experiments with kernel sizes of 1, 2, and 3 to show that a kernel size of $3\times1$ produces superior results at a reduced memory capacity.

\parag{Interpolation strategy}
The aforementioned position-aware linear interpolation mechanism maintains the non-uniform nature of samples. Simpler interpolation strategies such as nearest neighbour and average interpolation are substituted for this upsampling operation to show its merit. Nearest neighbour interpolation suffers the most in terms of perceptual quality while lending a 0.02s speedup in median time per iteration (0.57s) over our position-aware linear interpolation (0.59s). Experiments using averaged interpolation result in lower PSNR values than our interpolation strategy. Please refer to the supplementary material for a more detailed discussion. 

\subsection{Limitations and future work}
\label{sec: limitations}
As with NeRF, our method also struggles with modelling extremely thin structures through interpolation when samples along the visible surface are extremely sparse. UNeRF-Conv and UNeRF-Sub balance each other's strengths and weaknesses with UNeRF-Conv being more robust for dense scenes and UNeRF-Sub's resilience in the absence of diverse camera poses and limited data. Although we show that our UNeRF-Conv method converges in fewer epochs than NeRF at full capacity, UNeRF-Conv is still not fast enough for real-time applications. PyTorch has recently introduced a \textit{channels last} memory format~\cite{channelslastpytorch} which reports an 8-35\% performance gain for the memory layout we use and avoids dual permutations in our implementation. It is currently only implemented for 4D tensors, however if it is implemented for 3D inputs in the future, this bottleneck may be mitigated.

%% file: tables/tab_summary_synthetic_static_metrics.tex
\begin{table}[b]
  \caption{\textbf{Performance summary on the Blender dataset.} Images are rendered at $400\times400$ resolution. Both UNeRF variants deliver comparable or even better image quality with improved memory consumption. UNeRF-Conv presents a 12\% improvement in training time.}
  \vspace{\baselineskip}
  \label{tab: blender-overview-metrics}
  \centering
  \footnotesize
  \begin{tabular}{lccccc}
    \toprule
    & \multicolumn{2}{c}{Memory (GB) $\downarrow$} & \multicolumn{2}{c}{Time $\downarrow$} & PSNR (dB) $\uparrow$ \\
    \cmidrule(lr){2-3}
    \cmidrule(lr){4-5}
    \cmidrule(lr){6-6}
    Model & Forward & Max & Per Iteration (s) & To NeRF (h)$^1$ & Mean$^2$ \\
    \midrule
    NeRF & 11.27 & 11.76 & 0.42 & 35 & 31.33 \\
    NeRF-Mem & 9.06 & \textbf{9.47} & \textbf{0.34} & 27.50 & 31.16 \\
    UNeRF-Fine-Sub-Mem & 9.07 & 9.62 & 0.36 & - & 31.23 \\
    UNeRF-Conv & 9.06 & 9.73 & 0.59 & 31.14 & \textbf{32.26} \\
    \bottomrule%
    \multicolumn{6}{l}{\scriptsize $^1$ Time to achieve a VAL PSNR of 34.29dB on the chair dataset.} \\
    \multicolumn{6}{l}{\scriptsize $^2$ Average Test PSNR across 7 Blender scenes.} \\
  \end{tabular}
\end{table}

%% file: figures/anerf.tex
\begin{figure}
    \centering
    \begin{subfigure}[b]{0.24\textwidth}
        \includegraphics[width=\textwidth,trim={0 1cm 2.5cm 1cm}, clip]{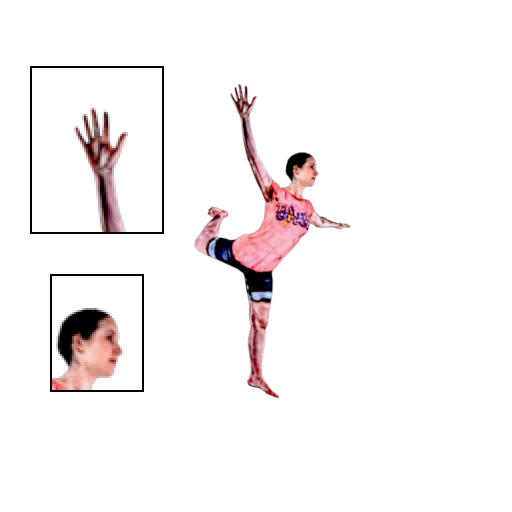}
        \caption{Ground truth}
        \label{fig:anerf-gt}
    \end{subfigure}
    \hfill
    \begin{subfigure}[b]{0.24\textwidth}
        \includegraphics[width=\textwidth, trim={0 1cm 2.5cm 1cm},clip]{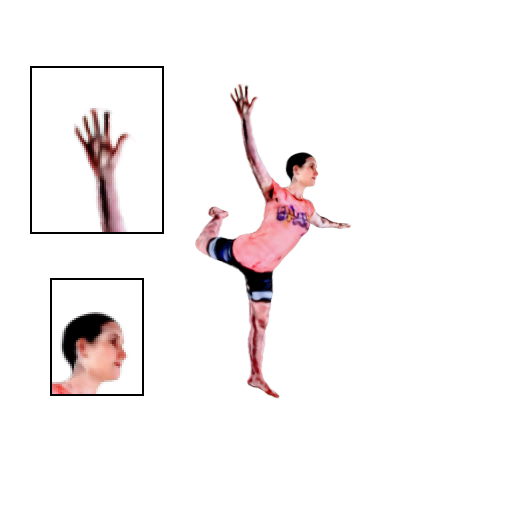}
        \caption{NeRF}
        \label{fig:anerf-nerf}
    \end{subfigure}
    \hfill
    \begin{subfigure}[b]{0.24\textwidth}
        \includegraphics[width=\textwidth, trim={0 1cm 2.5cm 1cm},clip]{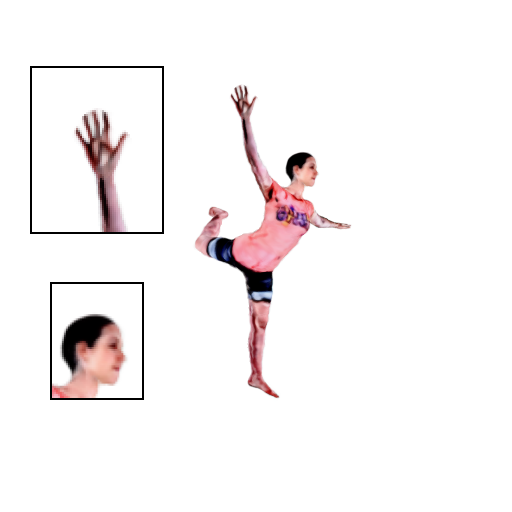}
        \caption{UNeRF-Sub}
        \label{fig:anerf-unerfdrop}
    \end{subfigure}
    \hfill
    \begin{subfigure}[b]{0.24\textwidth}
        \includegraphics[width=\textwidth, trim={0 1cm 2.5cm 1cm},clip]{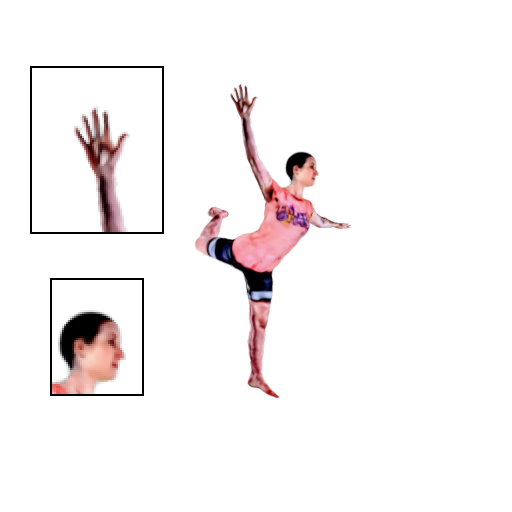}
        \caption{UNeRF-Conv}
        \label{fig:anerf-unerfconv}
    \end{subfigure}
    \caption{\textbf{Dynamic human motion rendering on the SURREAL dataset.} The 3D human pose and view are both unseen during training time. UNeRF-Conv preserves better finger structure  than the other architectures and achieves better results than the full capcity NeRF baseline when fully trained.}\label{fig:anerf}
\end{figure}

%% file: tables/tab_summary_anerf.tex
\begin{table}[htp]
  \caption{\textbf{Quantitative comparisons on rendering dynamic human motion from the SURREAL dataset.} Images are rendered at $512\times512$ resolution. UNeRF-Conv yields better image quality as well as memory consumption.}
  \vspace{\baselineskip}
  \label{table: anerf-overview-metrics}
  \centering
  \footnotesize
  \begin{tabular}{lcccccc}
    \toprule
     & & \multicolumn{2}{c}{Overall} & \multicolumn{2}{c}{Foreground} \\
    \cmidrule(lr){3-4}
    \cmidrule(lr){5-6}
    Architecture & Max Memory (GB) $\downarrow$ & PSNR $\uparrow$ & SSIM $\uparrow$ & PSNR $\uparrow$ & SSIM $\uparrow$ \\
    \midrule
    NeRF & 6.270 & 34.58 & 0.986 & 25.25 & 0.889 \\
    UNeRF-Sub & 5.221 & 33.94 & 0.984 & 24.62 & 0.875 \\
    UNeRF-Conv & 5.235 & \textbf{34.96} & \textbf{0.987} & \textbf{25.53} & \textbf{0.897} \\
    \bottomrule
  \end{tabular}
\end{table}

%% file: tables/tab_ablation_conv.tex
\begin{table}
  \caption{\textbf{Ablation study on convolution kernel size on the Blender dataset.} Images are rendered at $400\times400$ resolution. The larger UNeRF-Conv kernel size improves rendering outcomes.}
  \label{tab: unerf-conv-ablation}
  \vspace{\baselineskip}
  \footnotesize
  \centering
  \begin{tabular}{lcccccccccc}
    \toprule 
    & & \multicolumn{7}{c}{PSNR $\uparrow$} \\
    \cmidrule(lr){3-10}
    Model & Kernel & Chair & Drums & Ficus & Hotdog & Lego & Materials & Ship & Mean \\
    \midrule
    NeRF & - & 35.39 & 25.62 & 29.94 & 36.57 & 32.36 & 29.81 & 29.65 & 31.33 \\
    \cmidrule(lr){1-10}
    UNeRF-Conv & 3x1 & \textbf{35.69} & \textbf{26.11} & \textbf{31.63} & \textbf{37.75} & \textbf{33.30} & \textbf{31.73} & 29.62 & \textbf{32.26} \\
    UNeRF-Conv & 2x1 & 35.53 & 25.68 & 31.08 & 37.29 & 33.04 & 31.60 & \textbf{30.47} & 32.10 \\
    UNeRF-Conv & 1x1 & 34.90 & 25.41 & 29.58 & 36.26 & 31.59 & 29.71 & 29.62 & 31.01 \\
    \cmidrule(lr){1-10}
    UNeRF-Sub & - & 34.90 & 25.36 & 29.65 & 36.58 & 31.31 & 29.78 & 29.60 & 31.03 \\
    \bottomrule
  \end{tabular}
 \end{table}

%% file: tex/conclusion.tex
\section{Conclusion}
We present a method for training photorealistic scene rendering applications with a reduced optimization time and a significant reduction in the memory footprint compared to state-of-the-art methods while achieving higher fidelity results. Our UNeRF method with subsampling processes samples sparsely yet independently, allowing for it to achieve comparable results to NeRF especially on volumetrically sparse scenes with limited data and camera direction diversity. Although UNeRF-Conv shares information between adjacent samples along a ray, its restrictive neighbourhood allows for the method to disentangle geometry and view so long a sufficient number of training views is available. We show that our UNeRF-Conv method outperforms NeRF with respect to visual quality while reducing the memory footprint and by producing faster optimization times on both dynamic and static synthetic scenes. To the best of our knowledge, this is the first neural radience fields optimization method to report such an improvement in both memory and training time consumption while retaining its applicability to both static and dynamic scenes as well as presenting no image quality degradation. This is an important step towards training both time and memory conscious networks for photorealistic novel scene reconstruction.

\paragraph{Impact} 
\label{sec: impact} Even with our proposed method, training NeRF and related neural radiance fields models on GPUs may span several hours if not days, imposing significant environmental harm. We acknowledge the carbon footprint of training such work and encourage further research in the domain of reducing the resource demands of such methods. As memory availability can be a crucial bottleneck for training such methods, we hope the memory savings introduced in this work increase access for researchers with limited resources. 